\documentclass[letterpaper]{article} % DO NOT CHANGE THIS
\usepackage{aaai2026}  % DO NOT CHANGE THIS
\usepackage{times}  % DO NOT CHANGE THIS
\usepackage{helvet}  % DO NOT CHANGE THIS
\usepackage{courier}  % DO NOT CHANGE THIS
\usepackage[hyphens]{url}  % DO NOT CHANGE THIS
\usepackage{graphicx} % DO NOT CHANGE THIS
\usepackage{natbib}  % DO NOT CHANGE THIS AND DO NOT ADD ANY OPTIONS TO IT
\usepackage{caption} % DO NOT CHANGE THIS AND DO NOT ADD ANY OPTIONS TO IT
\usepackage{algorithm}
\usepackage{algorithmic}
\usepackage{amsmath}
\usepackage{booktabs}
\usepackage{bm}
\usepackage{pifont}
\usepackage{graphicx} 
\usepackage{epsfig}
\usepackage{epstopdf}

\usepackage{newfloat}
\usepackage{listings}
\DeclareCaptionStyle{ruled}{labelfont=normalfont,labelsep=colon,strut=off} % DO NOT CHANGE THIS
\floatstyle{ruled}
\newfloat{listing}{tb}{lst}{}
\floatname{listing}{Listing}
\title{Redundancy-optimized Multi-head Attention Networks 

for Multi-View Multi-Label Feature Selection}
\author{
    Yuzhou Liu\textsuperscript{\rm 1, \rm 2},
    Jiarui Liu\textsuperscript{\rm 1, \rm 2},
    Wanfu Gao\textsuperscript{\rm 1, \rm 2}\thanks{Corresponding author}
}
\affiliations{
    \textsuperscript{\rm 1} College of Computer Science and Technology, Jilin University, China\\
    \textsuperscript{\rm 2} Key Laboratory of Symbolic Computation and Knowledge Engineering of Ministry of Education, Jilin University, China\\
    liuyuzhou@jlu.edu.cn,
    liujr24@mails.jlu.edu.cn,
    gaowf@jlu.edu.cn
}

\begin{document}

\maketitle

\begin{abstract}
Multi-view multi-label data offers richer perspectives for artificial intelligence, but simultaneously presents significant challenges for feature selection due to the inherent complexity of interrelations among features, views and labels. Attention mechanisms provide an effective way for analyzing these intricate relationships. They can compute importance weights for information by aggregating correlations between Query and Key matrices to focus on pertinent values. However, existing attention-based feature selection methods predominantly focus on intra-view relationships, neglecting the complementarity of inter-view features and the critical feature-label correlations. Moreover, they often fail to account for feature redundancy, potentially leading to suboptimal feature subsets. To overcome these limitations, we propose a novel method based on \textbf{R}edundancy-optimized \textbf{M}ulti-head \textbf{A}ttention \textbf{N}etworks for \textbf{M}ulti-view \textbf{M}ulti-label \textbf{F}eature \textbf{S}election (RMAN-MMFS). Specifically, we employ each individual attention head to model intra-view feature relationships and use the cross-attention mechanisms between different heads to capture inter-view feature complementarity. Furthermore, we design static and dynamic feature redundancy terms: the static term mitigates redundancy within each view, while the dynamic term explicitly models redundancy between unselected and selected features across the entire selection process, thereby promoting feature compactness. Comprehensive evaluations on six real-world datasets, compared against six multi-view multi-label feature selection methods, demonstrate the superior performance of the proposed method.
\end{abstract}


\section{Introduction}

The increasing prevalence of multi-modal data acquisition has led to the widespread use of multi-view multi-label data, where samples are described by multiple feature views and associated with multiple labels simultaneously \cite{a2, a3, a5}. For example, an image can be represented by various visual descriptors such as HOG, color histograms and SIFT features, while being annotated with multiple tags like ``sky", ``river" and ``desert" \cite{b3, a6}. Feature selection for such data aims to identify a discriminative and non-redundant feature subset by leveraging complementary information across views while balancing the relevance between features and multiple labels. This process is crucial to improving the performance of the model \cite{c3}. Consequently, multi-view multi-label learning has gained significant traction in addressing complex real-world classification problems within domains such as machine learning and computer vision \cite{a4, b2}.
\begin{figure}[t]
\centering
\includegraphics[width=1.0\columnwidth]{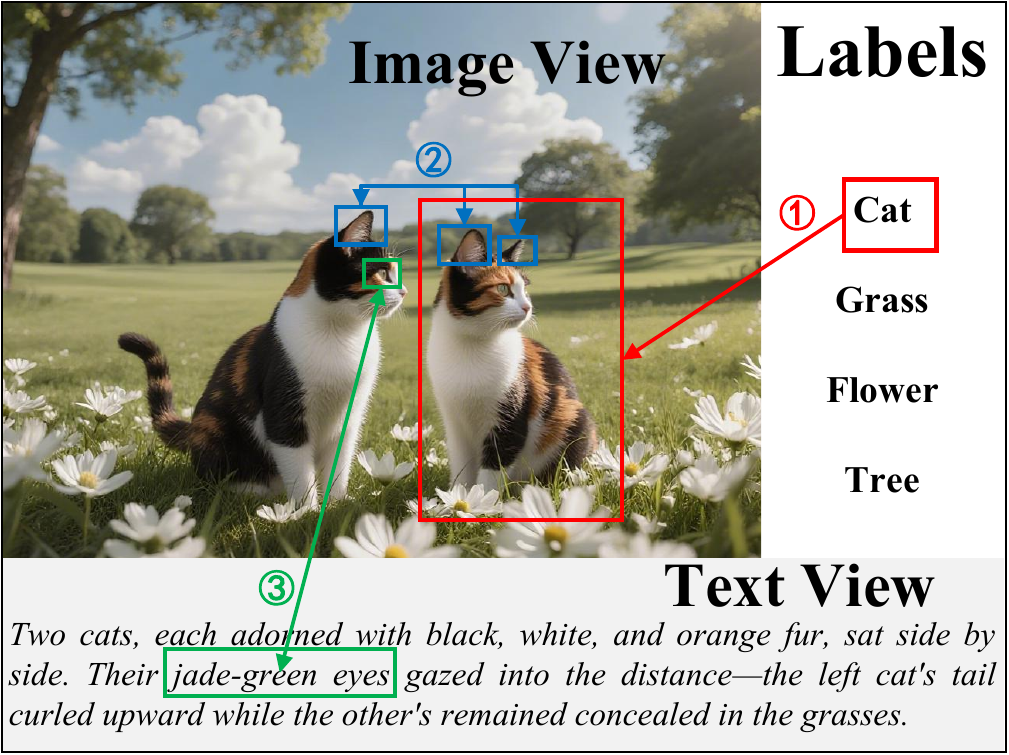}
\caption{ Example of relationships in MVML data. \ding{172} Feature-label correlations determine which features require attention. \ding{173} Inter-view relations may indicate redundant features (all of them are ``\textit{ear}”). \ding{174} Cross-view relations could show complementary features (the \textit{eyes} are ``\textit{jade-green}”).}
\label{fig2}
\end{figure}

Multi-view multi-label data typically exhibits high dimensionality with redundant and noisy features \cite{c5, b5}. More critically, the richness of its representation introduces intricate relationships. Figure 1 shows an example: (1) Correlations between features and labels influence selection results. For instance, the label “cat” would select features from the red box in the image view; (2) Features within the same view may exhibit higher redundancy due to shared origins, like features “ear” appearing many times in one image; (3) Features across different views can provide complementary information, for example, the feature “jade-green” in text view supplements the feature “eye” in image view. Effectively capturing these multifaceted relationships is essential to identify the most representative and discriminative features, yet it significantly complicates data storage, analysis, and application \cite{c6}.

Deep learning-based feature selection effectively mitigates the ``curse of dimensionality" \cite{c7}. Among these, attention mechanisms enable models to dynamically focus on the most informative components while suppressing irrelevant details, aligning naturally with feature selection goals by assigning importance weights to features, facilitating the identification of relevant features and the suppression of irrelevant ones \cite{c8}. However, current attention-based approaches for multi-view feature selection focus mainly on feature-label relationships within individual views. They largely overlook potential interactions between views and fail to adequately address feature redundancy, both intra-view and inter-view. This oversight result in highly redundant selected features, thus reducing the precision and efficiency of the feature selection process \cite{c9, c10}.

To address these critical gaps, we propose RMAN-MMFS (Redundancy-optimized Multi-head Attention Networks for Multi-view Multi-label Feature Selection). For the features-label relationships within views, our framework leverages multi-head attention, where each head specializes in analyzing feature-label relationships within a single view. To capture the complementarity of features between views, we incorporate cross-attention mechanisms between different heads to explicitly model complementary relationships among features across views. Furthermore, recognizing the detrimental impact of redundancy, we introduce dedicated static and dynamic redundancy optimization terms. The static term minimizes redundancy among features within each view, while the dynamic term actively reduces redundancy between unselected and already selected features during the iterative selection process. This combined strategy significantly enhances the quality and compactness of the selected feature subset. 

Extensive experiments conducted on six real-world datasets, comparing RMAN-MMFS against six representative multi-view multi-label feature selection baselines, validate its effectiveness and superiority. 

In summary, the main contributions of this paper can be summarized as follows:

\begin{itemize}
    \item A novel attention fusion framework: We integrate multi-head attention for intra-view analysis and cross-attention for inter-view complementarity modeling within a unified feature selection architecture.
    \item Comprehensive redundancy modeling: We design both static and dynamic redundancy optimization terms to effectively capture and minimize redundancy within views and between selected/unselected features throughout the selection process.
    \item Empirical validation: Rigorous experimentation on diverse real-world datasets demonstrates the superior performance of RMAN-MMFS over six existing state-of-the-art methods.
\end{itemize}

\section{Related Work}
\subsection{Multi-View Multi-Label Research}
Based on whether model training is involved, existing multi-view feature selection strategies can be broadly categorized into two types: (1) Non-training strategies (e.g. filter methods); (2) Training-integrated strategies (e.g. wrapper or embedded methods). The latter suffers from high time complexity, overfitting risks, and limited scalability to high-dimensional data, often converging to local optima. In contrast, non-training methods are the most widely used feature selection due to their low computational cost and robustness against overfitting, making them preferable for high-dimensional data. Method \cite{c19} normalizes mutual information in the mRMR criterion to mitigate the dominance of correlations or redundancy. Information-theoretic method like \cite{c20} integrates random variable distributions with granular computing, developing a algorithm based on mutual information and label enhancement. Methods \cite{c21, c22} leverage conditional mutual information to dynamically evaluate the relevance between selected features and candidates. They further propose novel label redundancy terms to better assess its effect on candidate feature relevance.

However, existing methods capture global view significance but neglect local importance, like inter-view complementarity or feature-specific relevance \cite {a1, b4}. Additionally, high dimensionality of multi-view data introduces optimization challenges due to increased parameters. To address these, attention-based feature selection has emerged.

\subsection{Attention-Based Multi-View Feature Selection}

Attention mechanisms have been used in common feature selection \cite{b6}. For example, \cite{c23} proposed a multi-head attention feature selector with $KQV$ correspond feature, label and a transformed feature matrix respectively. \cite{c24} employed an attention module under cognitive bias constraints. However, these methods are limited to single-view data and cannot address the complexities of multi-view multi-label scenarios.

Existing attention-based multi-view feature selection methods primarily adopt non-training strategies. Based on Formula (1), methods \cite{c25} utilize distinct attention networks to learn important features while suppressing noise. Method \cite{c26} employs multi-head attention networks to evaluate feature importance within domain-adaptive views. The MVFC method \cite{c27} denoises multi-view features via subspace learning. Cross-view strategies capture richer correlations and complementarity. For example, method \cite{c28} (MML-DAN) uses dual-attention networks to model deep interactions between label-specific views while balancing view-label relevance.

While these algorithms may achieve reasonable performance, they inadequately model complex data relationships. Furthermore, cross-view attention remain relatively scarce, and no existing method simultaneously considers intra-view feature relevance, inter-view feature complementarity and cross-feature redundancy. To address these limitations, we propose a cross-view attention-based approach that reduces information redundancy and ensures more comprehensive modeling of feature relationships.

\section{The Proposed Method}

\subsection{Definitions and Overall Method}

Mathematically, the attention mechanism can be defined as follows: for an input sequence $X=(x_1,x_2,...,x_n)$, the attention mechanism calculates a set of attention weights $W=(\omega_1,\omega_2,...,\omega_n)$, where each $\omega_i$ denotes the importance of the corresponding input element $x_i$. These weights are computed by a compatibility function $F$ between the query matrix $Q$ and the input $X$ , such as the scaled dot product. The unnormalized scores are normalized via the softmax function, as formalized in:
\begin{equation}
W=\mathrm{softmax}\left(\frac{Q^TK}{\sqrt{d_k}}\right).
\end{equation}

Here $Q$ is the query matrix, $K$ is the key vector for $X$, and $d_k$ denotes the dimensionality of $K$ for scaling stability. The resulting weights $W$ are then used to calculate the weighted sum of the input values, yielding the attention score $Y = WV$, where $V$ is the value vector for $X$. In essence:
\begin{equation}
Attention(K,Q,V)=\mathrm{softmax}\left(\frac{Q^TK}{\sqrt{d_k}}\right)V.
\end{equation}

Applying the attention mechanism is to feature selection lacks explicit $K$, $Q$ and $V$ matrices, requiring designed representations to capture the feature-label relationship for effective use.

Consider a multi-view dataset with $H$ views, denoted as $\{X^{(v)}\}_{v=1}^{H}$, where $X^{(v)}\in\mathcal{R}^{n\times d_v}$ is the feature matrix of the $v$-th view, $n$ is the sample count, and $d_v$ is its feature dimensionality. The multi-label matrix is $Y\in\mathcal{R}^{n\times c}$, with $c$ being the total label count.

Then, we map this to the attention framework: $K^{(v)}$ is the standardized feature matrix $X_{norm}^{(v)}$ for the $v$-th view, where $X_{norm}^{(v)}=\frac{X^{(v)}-\mu^{(v)}}{\sigma^{(v)}}$, $\mu^{(v)}$ and $\sigma^{(v)}$ are the mean and standard deviation vector of $X^{(v)}$ respectively. $Q$ is the label matrix $Y$, and $V$ is $X_{norm}$. Based on such matching, we design our approach's framework as shown in Figure 2.

\begin{figure*}[t]
\centering
\includegraphics[width=0.9\textwidth]{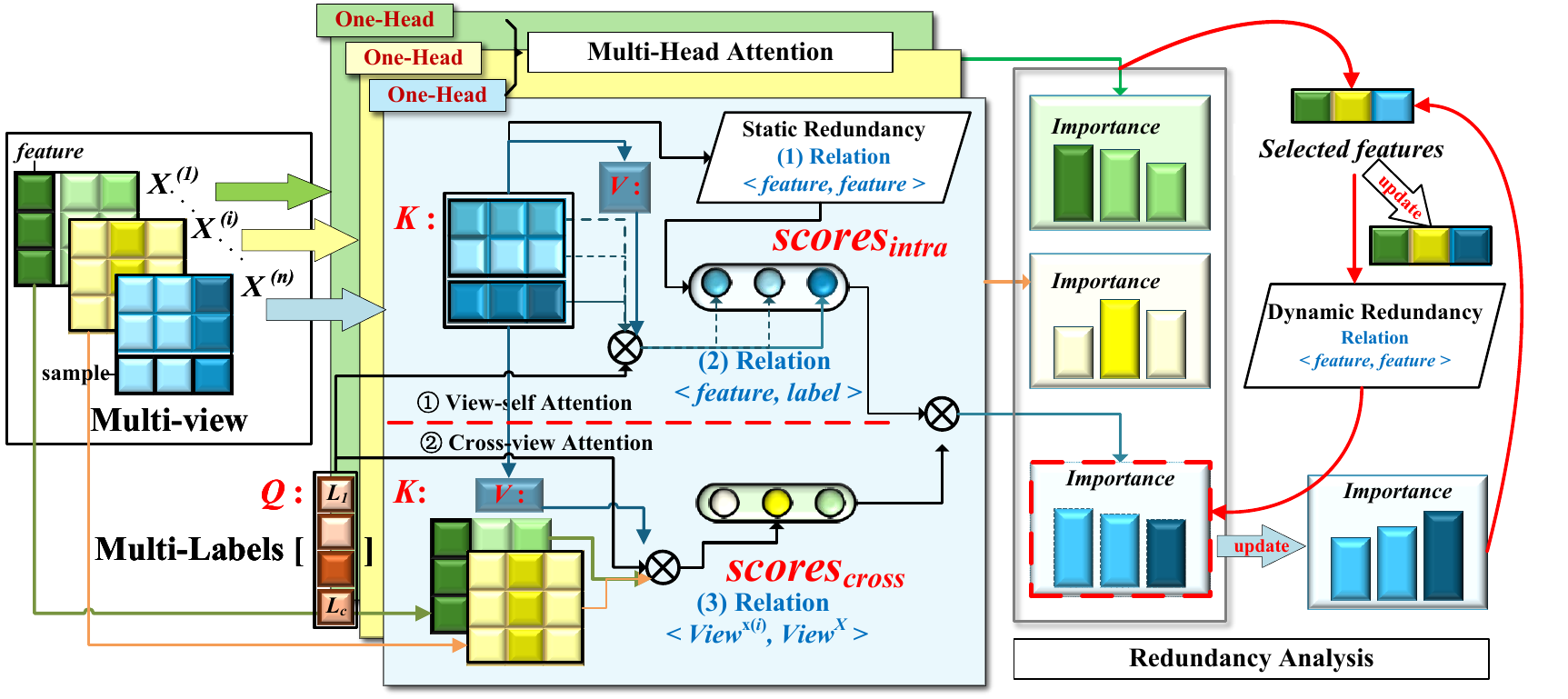} % Reduce the figure size so that it is slightly narrower than the column.
\caption{An illustration of the proposed RMAN-MMFS is presented, showcasing: (1) Multi-head attention for multi-view data analysis, which captures feature correlations and complementarity; (2) Static and dynamic redundancy penalty terms that quantify inter-feature redundancy. Feature weights are subsequently computed through this comprehensive integration process.}
\label{fig1}
\end{figure*}

% Using the \centering command instead of \begin{center} ... \end{center} will save space
% Positioning your figure at the top of the page will save space and make the paper more readable
% Using 0.95\columnwidth in conjunction with the

Overall, RMAN-MMFS comprises two parts: multi-head attention and redundancy optimization. In multi-head attention, each head computes: (1) View-self attention for feature-label relations within one view; (2) Cross attention for relations between the current view and others. Redundancy optimization also has two types: static redundancy considers relations among features within one view, while dynamic redundancy considers relations between candidate features across views. Details follow in this Section.
\subsection{Multi-Head Attention for Multi-View Data}
We apply multi-head attention to multi-view data by assigning one head per view. Each head independently learns view-specific feature-label and view-view interaction patterns.\\
\textbf{View-Self Attention.} For each view, we calculate its view-self attention, using the label matrix  $Querry$, where $Key$ and $Value$ are obtained from the features $X^{(v)}\in\mathcal{R}^{n\times d_v}$.  The resulting attention values become the initial importance scores $scores_{intra}^{(v)}$ for features in this view:\\
\begin{equation}
scores_{intra}^{(v)}=\mathrm{softmax}\left(\frac{Q^TK^{(v)}}{\sqrt{d_k}}\right)V^{(v)}.
\end{equation}

Each attention head corresponds to one view to avoid overfitting from direct raw data computation, yielding attention weights per view. Learning distinct attention patterns in separate subspaces enables model to capture richer information through multi-perspective feature modeling and independent parameterization for heterogeneous data distributions.

However, the core value of multi-view data is inter-view complementarity: different views provide relatively independent perspectives that characterize samples comprehensively, improving feature selection thoroughness. Lacking cross-view interaction leads to incomplete feature importance assessment. Thus, we introduce cross-view attention.\\
\textbf{Cross-Views Attention.} Cross-attention models relationships between distinct sequences or modalities.  Unlike self-attention, it enables one sequence to attend to another, proving valuable for integrating information from different sources, like machine translation and image captioning \cite{c11}.

To enable mutual attention across views: For the view $X^{(v)}$, features from all other views are concatenated into the context key matrix $K_{context}^{(v)}=Concat\left(K^{(1)},\ldots,K^{(v-1)},K^{(v+1)},\ldots,K^{(H)}\right)$, where $K_{context}^{(v)}\in\mathcal{R}^{n\times D_{context}}$, $D_{context}=\sum_{u\neq v}d_u$; The label-driven query matrix $Q$ interacts with $K$ to compute cross-view feature relevance:\\
\begin{equation}
\alpha_{cross}^{(v)}=\mathrm{~softmax}\left(\frac{Q^TK_{context}^{(v)}}{\sqrt{d_k}}\right)V^{(v)}.
\end{equation}

Mapping cross-view attention to the current view feature space:
\begin{equation}
scores_{cross}^{(v)}=\alpha_{cross}^{(v)}\cdot K_{context}^{(v)}\cdot X_{norm}^{(v)}.
\end{equation}

This dynamically quantifies inter-view complementarity through weight allocation. High weights indicate strong relevance between the current view and context features, facilitating synergistic information integration. This supplements view-self attention weights with cross-view feature complementarity, providing a robust global representation for redundancy penalization and feature selection.

\subsection{Redundancy Optimization}
As mentioned, solely the attention mechanism inadequately addresses feature redundancy. Thus, we introduce redundancy optimization terms. Typically, multi-view data redundancy manifests in two forms: (1) Intra-view redundancy: High correlation among features within the same view. (2) Global redundancy: Information overlap between candidate features and selected features. Redundant features reduce model generalizability and obscure discriminative features. To address this, we introduce static and dynamic redundancy optimizations.\\
\textbf{Static Redundancy Optimization.} This measures intra-view linear redundancy using the mean absolute correlation coefficient (MACC) between features. For each feature $i$ in current view, it is calculated as:\\
\begin{equation}
R_{static}^{(v)}(i)=\frac{1}{d_v-1}\sum_{i\neq j}\left|Corr\left(X_i^{(v)},X_j^{(v)}\right)\right|.
\end{equation}

This optimizes highly correlated features within view by statically adjusting the attention weights, enabling rapid intra-view redundancy reduction, and we use it to adjust the initial $scores_{intra}^{(v)}$ to get a more reasonable results.\\
\textbf{Dynamic Redundancy Optimization.} Maintains a global set $S$ tracking indices of selected features $f$ (view and position). For candidate feature $i$, its mutual information (MI) with all features in $S$ is computed:
\begin{equation}
R_{dynamic}^{(v)}(i)=\frac{1}{|S|}\Sigma_{i\in C}\sum_{f\in S}MI(i,f).
\end{equation}

The generated attention weights are dynamically adjustable, capturing nonlinear dependence between cross-view and global features while accurately suppressing global redundancy in candidate and selected features, avoiding invisible repeated selection of redundant features. Based on the dynamic redundancy optimization, the feature selection results are updated dynamically in an iterative way as shown in Figure 2.

The two redundant optimizations work in synergy. The static term rapidly eliminates intra-view redundancy, while the dynamic term refines global feature selection. Combining linear (MACC) and nonlinear (MI) metrics balances efficiency and precision. This design effectively mitigates redundancy in multi-view multi-label tasks, yielding compact, discriminative feature subsets for downstream classifiers.
\subsection{Objective Function}
Integrating multi-head attention and redundancy analysis, the unified objective function is:
\begin{equation}  % 整个公式只有一个编号
\begin{split}
Importance^{(v)} 
= & \bigg\| \bigg[ \Big( scores_{\text{intra}}^{(v)} - \lambda \cdot R_{\text{static}}^{(v)} \Big)\\ 
& + scores_{\text{cross}}^{(v)} \bigg] - \beta \cdot R_{\text{dynamic}}^{(v)} \bigg\|_2.
\end{split}
\label{eq:importance}
\end{equation}

Where $\lambda$ and $\beta$ are penalty coefficients. A detailed breakdown of this process is elucidated in Algorithm 1.

\begin{algorithm}[tb]
\caption{RMAN-MMFS}
\label{alg:mhcar}
\textbf{Input}: Data matrices $\{X^{(v)}\}_{v=1}^{H}$, Label matrix $Y$ \\
\textbf{Parameter}: Parameters $\lambda$ and $\beta$ \\
\textbf{Output}: Set of selected features $F$
\begin{algorithmic}[1]
\STATE Initialize $F \leftarrow \emptyset$ \COMMENT{Selected features set}
\STATE $X_{\text{norm}} \leftarrow \frac{X - \mu}{\sigma}$ \COMMENT{Normalize data}
\FOR{$v = 1$ \TO $H$}
    \STATE Define $Q \leftarrow Y$, $d_k \leftarrow \dim(X^{(v)})$ 
    \STATE Compute cross-view attention via Formula (4)
    \STATE Map to feature space via Formula (5)
    \STATE Define $K^{(v)} \leftarrow F(X^{(v)})$, $V^{(v)} \leftarrow X^{(v)}$
    \STATE Calculate view-self attention via Formula (3)
    
    \FOR{each feature $i$ in $X^{(v)}$}
        \STATE Compute $R_{\text{static},i}^{(v)}$ via Formula (6)
    \ENDFOR
    \STATE $scores_{\text{intra}}^{(v)} \leftarrow scores_{\text{intra}}^{(v)} - \lambda R_{\text{static}}^{(v)}$
    \STATE $scores^{(v)} \leftarrow scores_{\text{intra}}^{(v)} + scores_{\text{cross}}^{(v)}$
    
    \IF{$F \neq \emptyset$}
        \FOR{each feature $i$ in $X^{(v)}$}
            \STATE Compute $R_{\text{dynamic},i}^{(v)}$ via Formula (7)
        \ENDFOR
    \ENDIF
    \STATE $scores^{(v)} \leftarrow scores^{(v)} - \beta R_{\text{dynamic}}^{(v)}$
    \STATE $Importance^{(v)} \leftarrow \|scores^{(v)}\|_2$ \COMMENT{L2 norm}
    \STATE Sort $Importance^{(v)}$ descending $\rightarrow I_{\text{sorted}}$
    \STATE Add top features from $I_{\text{sorted}}$ to $F$
\ENDFOR
\STATE \textbf{return} $F$
\end{algorithmic}
\end{algorithm}

\section{Experiments}
\subsection{Datasets}
We evaluate our method on six public multi-view multi-label datasets: SCENE, VOC07, MIRFlickr, OBJECT, Yeast and Mfeat. The first four are real-world image datasets from \cite{c12}. Yeast contains gene expression data of yeast under different conditions, with each sample belonging to one of 14 categories. Mfeat includes six feature views of handwritten digits 0-9, with each sample belonging to one of ten categories. Table 1 details each dataset, including the number of views, features, samples and labels .
\begin{table}[t]
\centering
\setlength{\tabcolsep}{1.4pt} % 减小列间距
\begin{tabular}{l | c c c c c c}
    \toprule
    Views&SCENE&VOC07&MIRFlickr&OBJECT&Yeast&Mfeat\\
    \midrule
    View1&64&100&100&64&79&76\\
    View2&225&512&512&225&24&216\\
    View3&144&100&100&144&-&64\\
    View4&73&-&-&73&-&240\\
    View5&128&-&-&128&-&47\\
    View6&-&-&-&-&-&6\\
    \midrule
    Features&634&712&712&634&103&649\\
    Samples&4400&3817&4053&6047&2417&2000\\
    Labels&31&20&38&31&14&10\\
    \bottomrule
\end{tabular}
\caption{Description of datasets.}
\label{table1}
\end{table}
\subsection{Baseline Methods}
Information-theoretic and sparsity-based methods are mature and competitive feature selection baselines. Both information-theoretic methods and attention-based methods are lightweight and do not require training data. Therefore, we compare RMAN against three information-theoretic and three sparsity-based feature selection methods which are prominent: STFS \cite{c13}, ENM \cite{c14}, MLSMFS \cite{c16}, MSFS \cite{c12}, DHLI \cite{c30} and EF2FS \cite{c31}.

\subsection{Evaluation Metrics}
We selected four common metrics in this field for evaluation:

\begin{itemize}
\item Average Precision (AP): Measures precision across recall levels via the area under the precision-recall curve for each label.
\item Macro-Average Area Under Curve (AUC): Averages each label’s AUC value, evaluating the model’s positive-negative distinction across classification thresholds.
\item Coverage Error (CE): Measures the average extra labels needed to cover all true labels.
\item Ranking Loss (RL): Reflects the probability of irrelevant labels ranking above relevant ones in the predicted ranking across samples.
\end{itemize}

Higher values indicate better performance for AP and AUC, while lower values are better for CE and RL. Additionally, AP and AUC are label-based metrics, whereas CE and RL are sample-based, providing multidimensional evaluation \cite{b1,c18}.

\subsection{Experiment Settings}
Python was used as the programming language. We selected MLKNN (k=10) as the classifier. Feature selection ranged from 2\% to 20\% of total features, increasing by 2\% per iteration. In each experiment, 30\% of samples were for testing and 70\% for training. Experiments were repeated 10 times with averaged results. Hyperparameters for other methods were set according to their original papers. Multi-view datasets were concatenated for feature selection in other information-theoretic methods.

\subsection{Experimental Results and Analysis}
Tables 2-3 show method performance on six datasets using four metrics. The ``Average'' row displays mean results across all datasets. Bold values indicate the best classification performance on a specific dataset, while underlined values denote second-best. Up/down arrows signify higher/lower values indicate better performance.
\begin{table*}[t]
\centering
\small % 减小字体尺寸
\setlength{\tabcolsep}{7pt} % 减小列间距
\begin{tabular}{@{}l@{\hspace{2pt}} | c c c c c c c@{}}% 左对齐并压缩左边距
    \toprule
    Datasets& RMAN & MLSMFS & DHLI & STFS & MSFS & EF2FS & ENM \\
    \midrule
    \multicolumn{8}{c}{\textbf{AP} $\uparrow$} \\
    \midrule
			SCENE     & \textbf{0.260} $\pm$ \textbf{0.010} & 0.231 $\pm$ 0.013 & 0.244 $\pm$ 0.015& 0.229 $\pm$ 0.014 & 0.233 $\pm$ 0.010 & \underline{0.250 $\pm$ 0.006} & 0.227 $\pm$ 0.014 \\
			VOC07     & \textbf{0.136} $\pm$ \textbf{0.003} & 0.115 $\pm$ 0.002 & 0.128 $\pm$ 0.002 & \textbf{0.136} $\pm$ \textbf{0.007} & 0.111 $\pm$ 0.001 & 0.113 $\pm$ 0.006 & \underline{0.131 $\pm$ 0.005} \\
			Yeast   & \textbf{0.324} $\pm$ \textbf{0.006} & 0.316 $\pm$ 0.011 & 0.321 $\pm$ 0.006 & 0.314 $\pm$ 0.012 & 0.311 $\pm$ 0.003 & 0.315 $\pm$ 0.006 & \underline{0.323 $\pm$ 0.010} \\
			MIRFlickr & \textbf{0.296} $\pm$ \textbf{0.009} & 0.260 $\pm$ 0.001 & 0.254 $\pm$ 0.002 & 0.288 $\pm$ 0.012 & 0.243 $\pm$ 0.001 & 0.269 $\pm$ 0.037 & \underline{0.293 $\pm$ 0.011} \\
		  Mfeat     & 0.844 $\pm$ 0.100 & \textbf{0.944} $\pm$ \textbf{0.060} & 0.729 $\pm$ 0.075 & \underline{0.891 $\pm$ 0.082} & 0.658 $\pm$ 0.151 & 0.780 $\pm$ 0.032 & 0.651 $\pm$ 0.131 \\
			OBJECT    & \textbf{0.128} $\pm$ \textbf{0.018} & 0.098 $\pm$ 0.011 & 0.117 $\pm$ 0.019 & 0.104 $\pm$ 0.011 & 0.107 $\pm$ 0.009 & \underline{0.125 $\pm$ 0.010} & 0.102 $\pm$ 0.013 \\
			Average   & \textbf{0.331} & \underline{0.327} & 0.299 & \underline{0.327} & 0.278 & 0.309 & 0.288 \\
			\midrule
			
			\multicolumn{8}{c}{\textbf{AUC} $\uparrow$} \\
			\midrule
			SCENE     & \textbf{0.658} $\pm$ \textbf{0.016} & 0.596 $\pm$ 0.023 & \underline{0.622 $\pm$ 0.032} & 0.613 $\pm$ 0.037 & 0.593 $\pm$ 0.023 & 0.607 $\pm$ 0.013 & 0.603 $\pm$ 0.033 \\
			VOC07     & \textbf{0.612} $\pm$ \textbf{0.020} & 0.536 $\pm$ 0.005 & 0.575 $\pm$ 0.008 & \underline{0.609 $\pm$ 0.021} & 0.500 $\pm$ 0.001 & 0.505 $\pm$ 0.008 & 0.595 $\pm$ 0.020 \\
			Yeast     & \underline{0.541 $\pm$ 0.015} & \textbf{0.542} $\pm$ \textbf{0.018} & 0.536 $\pm$ 0.016 & 0.536 $\pm$ 0.020 & 0.512 $\pm$ 0.007 & 0.526 $\pm$ 0.010 & 0.540 $\pm$ 0.021 \\
			MIRFlickr & \textbf{0.618} $\pm$ \textbf{0.017} & 0.550 $\pm$ 0.001 & 0.546 $\pm$ 0.005 & 0.603 $\pm$ 0.015 & 0.509 $\pm$ 0.001 & 0.564 $\pm$ 0.009 & \underline{0.613 $\pm$ 0.020} \\
			Mfeat     & 0.965 $\pm$ 0.020 & \textbf{0.985} $\pm$ \textbf{0.012} & 0.878 $\pm$ 0.026 & \underline{0.978 $\pm$ 0.014} & 0.973 $\pm$ 0.057 & 0.623 $\pm$ 0.034 & 0.909 $\pm$ 0.048 \\
			OBJECT    & \textbf{0.653} $\pm$ \textbf{0.030} & 0.610 $\pm$ 0.023 & \underline{0.634 $\pm$ 0.033} & 0.619 $\pm$ 0.028 & 0.595 $\pm$ 0.019 & \underline{0.634 $\pm$ 0.011} & 0.619 $\pm$ 0.030 \\
			Average   & \textbf{0.675} & 0.637 & 0.632 & \underline{0.660} & 0.612 & 0.577 & 0.647 \\
    \bottomrule
\end{tabular}
\caption{Experimental results of all methods in terms of AP and AUC (mean $\pm$ std).}
\label{table2}
\end{table*}

\begin{table*}[t]
\centering
\small % 减小字体尺寸
\setlength{\tabcolsep}{4.7pt} % 减小列间距
\begin{tabular}{@{}l@{\hspace{2pt}} | c c c c c c c@{}}% 左对齐并压缩左边距
    \toprule
    Datasets& RMAN & MLSMFS & DHLI & STFS & MSFS & EF2FS & ENM \\
    \midrule
    \multicolumn{8}{c}{\textbf{CE} $\downarrow$} \\
    \midrule
			SCENE     & \textbf{13.877} $\pm$ \textbf{0.253} & 14.786 $\pm$ 0.350 & 14.361 $\pm$ 0.468 & 14.806 $\pm$ 0.589 & 14.967 $\pm$ 0.490 & \underline{13.906 $\pm$ 0.230} & 14.792 $\pm$ 0.506 \\
			VOC07     & \textbf{8.332} $\pm$ \textbf{0.270} & 9.445 $\pm$ 0.044 & 8.744 $\pm$ 0.106 & \underline{8.443 $\pm$ 0.284} & 10.489 $\pm$ 0.437 & 10.481 $\pm$ 0.246 & 8.546 $\pm$ 0.275 \\
			Yeast     & \textbf{8.786} $\pm$ \textbf{0.142} & \underline{8.798 $\pm$ 0.353} & 8.997 $\pm$ 0.274 & 8.873 $\pm$ 0.347 & 9.298 $\pm$ 0.351 & 9.076 $\pm$ 0.172 & 8.878 $\pm$ 0.273 \\
			MIRFlickr & \textbf{21.575} $\pm$ \textbf{0.296} & 22.834 $\pm$ 0.020 & 23.318 $\pm$ 0.279 & 21.884 $\pm$ 0.210 & 25.701 $\pm$ 1.283 & 22.940 $\pm$ 0.365 & \underline{21.641 $\pm$ 0.383} \\
			Mfeat     & 1.418 $\pm$ 0.180 & \textbf{1.221} $\pm$ \textbf{0.153} & 2.210 $\pm$ 0.247 & \underline{$1.288 \pm 0.155$} & 2.443 $\pm$ 0.488 & 4.779 $\pm$ 0.291 & 1.959 $\pm$ 0.412 \\
			OBJECT    & \textbf{9.353} $\pm$ \textbf{0.500} & 10.092$\pm$ 0.354 & 9.728 $\pm$ 0.621 & 9.803 $\pm$ 0.441 & 10.292 $\pm$ 0.219 & 9.664 $\pm$ 0.201 & \underline{9.564 $\pm$ 0.554} \\
			Average   & \textbf{10.572} & 11.196 & 11.226 & 12.172 & 12.918 & 11.808 & \underline{10.897} \\
			\midrule
			\multicolumn{8}{c}{\textbf{RL} $\downarrow$} \\
			\midrule
			SCENE     & \textbf{0.090} $\pm$ \textbf{0.004} & 0.104 $\pm$ 0.005 & 0.097 $\pm$ 0.006 & 0.103 $\pm$ 0.008 & 0.108 $\pm$ 0.007 & \underline{0.091 $\pm$ 0.004} & 0.103 $\pm$ 0.007 \\
			VOC07     & \textbf{0.192} $\pm$ \textbf{0.009} & 0.231 $\pm$ 0.002 & 0.207 $\pm$ 0.003 & \underline{0.197 $\pm$ 0.010} & 0.267 $\pm$ 0.017 & 0.266 $\pm$ 0.010 & 0.201 $\pm$ 0.009 \\
			Yeast     & \textbf{0.250} $\pm$ \textbf{0.009} & 0.258 $\pm$ 0.011 & 0.257 $\pm$ 0.013 & 0.254 $\pm$ 0.015 & 0.274 $\pm$ 0.028 & 0.261 $\pm$ 0.009 & \underline{0.252 $\pm$ 0.011} \\
			MIRFlickr & \textbf{0.152} $\pm$ \textbf{0.006} & 0.178 $\pm$ 0.001 & 0.178 $\pm$ 0.003 & 0.160 $\pm$ 0.006 & 0.230 $\pm$ 0.039 & 0.181 $\pm$ 0.008 & \underline{0.153 $\pm$ 0.007} \\
			Mfeat     & 0.046 $\pm$ 0.022 & \textbf{0.025} $\pm$ \textbf{0.017} & 0.134 $\pm$ 0.027 & \underline{0.032 $\pm$ 0.017} & 0.160 $\pm$ 0.054 & 0.420 $\pm$ 0.032 & 0.107 $\pm$ 0.046 \\
			OBJECT    & \textbf{0.175} $\pm$ \textbf{0.014} & 0.199 $\pm$ 0.008 & 0.187 $\pm$ 0.017 & 0.187 $\pm$ 0.012 & 0.204 $\pm$ 0.006 & \underline{0.182 $\pm$ 0.005} & 0.187 $\pm$ 0.015 \\
			Average   & \textbf{0.151} & 0.166 & 0.177 & \underline{0.156} & 0.207 & 0.234 & 0.167 \\
    \bottomrule
\end{tabular}
\caption{Experimental results of all methods in terms of CE and RL (mean $\pm$ std).}
\label{table3}
\end{table*}
Our method achieves the highest performance on five datasets (SCENE, VOC07, Yeast, OBJECT and MIRFlickr) due to RMAN's comprehensive relationship modeling. However, RMAN underperforms MLSMFS on Mfeat. Mfeat exhibits significantly higher feature-label correlations than others, causing greater information overlap between features. This overlap complicates redundancy avoidance, weakening optimization adjustment and causing excessive penalization of discriminative features. Nevertheless, considering overall performance across all six datasets, RMAN outperforms others on AP, AUC, CE, and RL. Figure 3 further presents one dataset performance across all evaluation metrics to clearly demonstrate our results.
\begin{figure}[t]
\centering
\includegraphics[width=1.0\columnwidth]{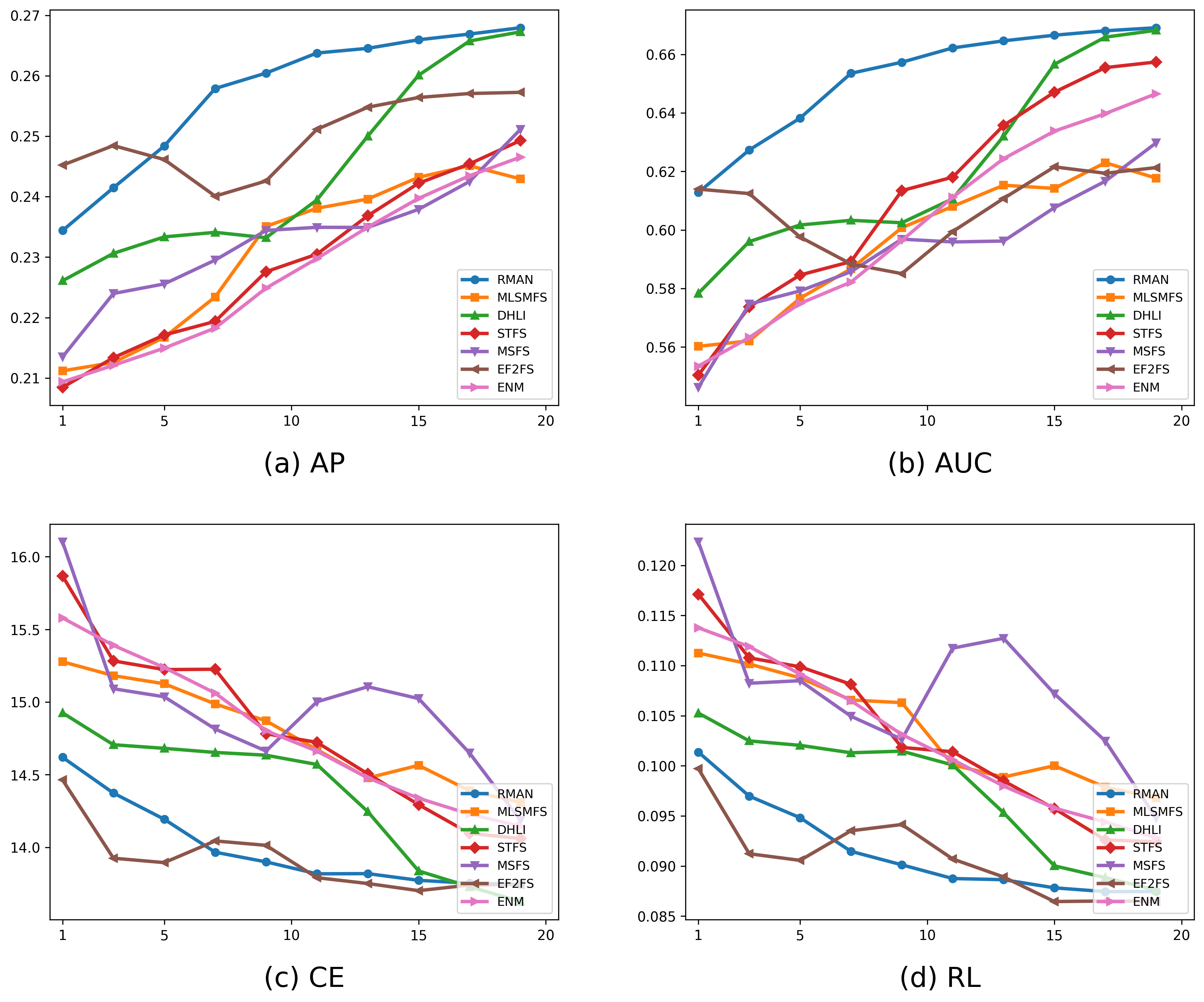} % Reduce the figure size so that it is slightly narrower than the column. Don't use precise values for figure width.This setup will avoid overfull boxes.
\caption{ Seven methods on SCENE in terms of AP, AUC, CE and RL.}
\label{fig2}
\end{figure}
\subsection{Ablation Experiments}
\textbf{Ablation Study} To further validate the effectiveness of each component, we conduct ablation experiments by removing cross-attention weight calculation and redundancy optimization terms from the objective function. Since view-self attention forms the architectural foundation, it is not ablated. This result in Formula (9) without cross-attention ($RMAN_1$):
\begin{equation}
\left\|\left(scores_{intra}^{(v)}-\lambda\cdot R_{static}^{(v)}\right)-\beta\cdot R_{dynamic}^{(v)}\right\|_2,
\end{equation}
Formula (10) without static redundancy penalty ($RMAN_2$):
\begin{equation}
\left\|\left[scores_{intra}^{(v)}+scores_{cross}^{(v)}\right]-\beta\cdot R_{dynamic}^{(v)}\right\|_2,
\end{equation}
and Formula (11) without dynamic redundancy penalty ($RMAN_3$):
\begin{equation}
\left\|\left(scores_{intra}^{(v)}-\lambda\cdot R_{static}^{(v)}\right)+scores_{cross}^{(v)}\right\|_2.
\end{equation}

These formulations were evaluated across six datasets.

\begin{table*}[t]
\centering
%\resizebox{.95\columnwidth}{!}{
\begin{tabular}{l c c c c | c c c c c c}
    \toprule
    &VSA& CRA & SRP & DRP &SCENE& VOC07 & Yeast & MIRFlickr & Mfeat & OBJECT\\
    \midrule
			RMAN&$\surd$ & $\surd$ & $\surd$ & $\surd$ & \textbf{0.260}  & \textbf{0.136} & \textbf{0.324} & \textbf{0.296} & \textbf{0.844}& \textbf{0.128} \\
			RMAN$_{1}$&$\surd$ &  & $\surd$ & $\surd$ &0.257 & 0.133 & 0.314 & 0.294 & 0.830 & 0.124  \\
			RMAN$_{2}$&$\surd$ & $\surd$ &  & $\surd$ &0.259 & 0.134 & 0.314 & 0.295 & 0.842 & 0.119 \\
			RMAN$_{3}$&$\surd$ & $\surd$ & $\surd$ & &0.256 & 0.133 & 0.316 & 0.294 & 0.825 & 0.123   \\
    \bottomrule
\end{tabular}
\caption{Ablation experimental results of RMAN on six datasets.}
\label{table1}
\end{table*}

Table 4 shows components in in each variant and AP metrics for ablation methods. VSA, CVA, SRP, and DRP represent view-self attention, cross-view attention, static redundancy term and dynamic redundancy term, respectively. Results show removing any component harms performance, highlighting their importance. Our full method achieves optimal performance on six datasets, proving its effectiveness.
\\
\textbf{Redundancy Optimization Design}
Feature selection aims to identify the most relevant and least redundant feature subset. Redundancy penalties reduce inter-feature redundancy. Our algorithm uses : (1) correlation coefficients (2) mutual information. To validate this design, we test different combinations for static/dynamic penalties. Table 5 shows CE metrics under different combinations. Results confirm our current configuration achieves optimal performance.
\begin{table}[t]
\centering
%\resizebox{.95\columnwidth}{!}{
    \begin{tabular}{l c c | c c c }
    \toprule
    & SRP & DRP & SCENE & VOC07 & Yeast \\
			\midrule
			RMAN&Corr&MI & \textbf{13.877} & \textbf{8.332} & \textbf{8.786} \\
			RMAN-$\alpha$&MI&Corr& 13.996 & 8.424 & 8.977 \\
			RMAN-$\beta$&MI&MI & 13.934 & 8.428 & 8.980 \\
			RMAN-$\gamma$&Corr&Corr& 13.956 & 8.441  & 9.018 \\
    \bottomrule
    \end{tabular}
\caption{The impact of different redundant computations on the CE metrics on three datasets.}
\label{table5}
\end{table}

\subsection{Parameter Analysis}
Figure 4 shows sensitivity analysis for two parameters on MIRFlickr. Optimize the grid by fixing other parameters. The X-axis represents grid search range (0.001 to 1000), Y-axis shows number of selected features, and Z-axis indicates AUC metric values. For parameter $\alpha$ and $\beta$, as the number of features increases, the indicator value rises slightly and tends to stabilize. Performance remains stable across feature subsets on MIRFlickr.
\begin{figure}[t]
\centering
\includegraphics[width=1.0\columnwidth]{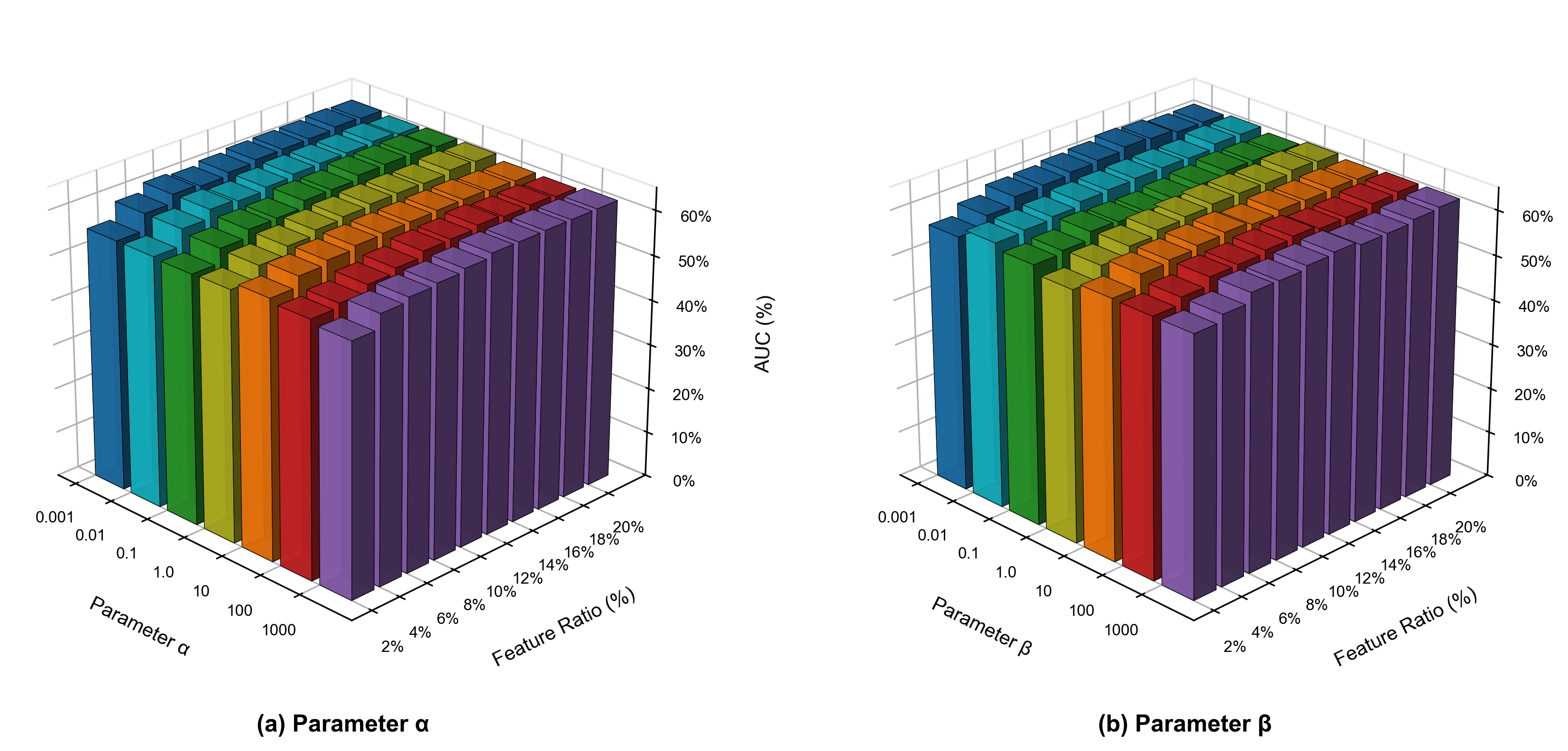} % Reduce the figure size so that it is slightly narrower than the column. Don't use precise values for figure width.This setup will avoid overfull boxes.
\caption{Parameter sensitivity studies on the SCENE datasets.}
\label{fig3}
\end{figure}

\subsection{Time Complexity of Methods}
We conducted complexity analysis of RMAN and performed comparative assessments against three training-data-free information-theoretic baselines. Complexity is determined by sample size ($n$), feature size ($d$), label count ($q$) and selected feature count ($k$). For RMAN, multi-head attention and cross-attention both have $O(nqd)$ complexity. Static and dynamic redundancy penalties have $O(d^2)$ and $O(ndk)$ complexity respectively. Thus RMAN's total complexity is $O(nqd + d^2 + ndk)$. Similarly, STFS, ENM and MLSMSF have complexities $O(ndq^2+kndq+ndq)$, $O(ndq)$ and $O(ndq^2+ndk+ndq)$ respectively. RMAN and ENM are cubic complexity, while the other two involve quartic terms, indicating our method's computational advantage. Actual runtimes on six datasets are shown in Table 6. Compared to ENM, RMAN exhibits marginally longer runtime, but RMAN delivers superior overall performance, demonstrating its ability to maintain optimal performance with lightweight computation.

\begin{table}[htb]
\centering
\small % 减小字体尺寸
\setlength{\tabcolsep}{2.3pt} % 减小列间距
%\resizebox{.95\columnwidth}{!}{
    \begin{tabular}{l | c c c c c c}
    \toprule
    Methods& SCENE & VOC07 & Yeast & MIRFlickr & Mfeat & OBJECT \\
    \midrule
			ENM& 558 & 345 & 123 & 616 & 137 & 659 \\
			MLSMFS& 26911 & 13015 & 562 & 36103 & 1622 & 4532 \\
			STFS& 215944 & 87146 & 742 & 207207 & 18470 & 301418 \\
			RMAN& 2502 & 2150 & 200 & 2358 & 2047 & 2868 \\
    \bottomrule
    \end{tabular}
\caption{The running time (in seconds).}
\label{table1}
\end{table}

\section{Conclusion}
This paper proposes a novel redundancy-optimized multi-head attention networks for multi-view multi-label feature selection, named RMAN-MMFS. Previous attention-based feature selection methods fail to adequately consider inter-view feature complementarity and overlook the guiding role of feature redundancy. To address these issues, RMAN-MMFS introduces multi-head attention to compute view-self feature-label relevance, employs cross-attention to handle inter-view feature complementarity, and designs both static and dynamic redundancy penalty terms. Experimental results demonstrate that our method outperforms existing state-of-the-art approaches across multiple aspects.
Our research improves feature selection accuracy by capturing relationships among features, views, and labels as comprehensively as possible. Future work will focus on exploring relationships between labels to further enhance the method's performance.

\section{Acknowledgments}
This work is funded by: Jilin Provincial Science and Technology Development Plan Project No.20240302084GX, and Changchun Scienceand Technology Bureau Project 23YQ05.

\bibliography{aaai2026}

\end{document}